\documentclass[11pt]{article}

\usepackage[final]{acl}
\usepackage{graphicx}
\usepackage{microtype}
\usepackage{times}
\usepackage{latexsym}
\usepackage{booktabs}
\usepackage{enumitem}
\usepackage{amsmath}
\usepackage{multirow}
\usepackage{multicol}
\usepackage{svg}
\usepackage[english,bidi=default]{babel} 
\babelfont{rm}{TeXGyreTermesX} 
\babelprovide[import]{hindi}
\babelfont[*devanagari]{rm}{Lohit Devanagari}
\babelprovide[import]{arabic}
\babelfont[*arabic]{rm}{Noto Sans Arabic}

\title{Is EEG-to-Text Feasible in Real-World Scenarios? An In-Depth Analysis Using a Neuropsychology-Inspired Benchmark}

\author{Zihan Zhang$^{1}$\thanks{Equal Contribution.}, Yu Bao$^{1,2}$\footnotemark[1], Xiao Ding$^1$\thanks{Corresponding authors.}, Tianyi Jiang$^3$, Kai Xiong$^4$ \\
  $^1$Research Center for Social Computing and Interactive Robotics, Harbin Institute of Technology  \\
  $^2$Shanghai Innovation Institute, Shanghai, China \\
  $^3$State Key Laboratory for Novel Software Technology, Nanjing University\\
  $^4$Zhongguancun Laboratory, Beijing, China \\
  \texttt{\{zihanzhang, ybao, xding\}@ir.hit.edu.cn, xiongk@zgclab.edu.cn} \\}

\begin{document}

\maketitle
\begin{abstract}

Translating brain signals into text could restore communication for people with severe paralysis, yet practically usable systems to date rely on invasive electrocorticography (ECoG). Electroencephalography (EEG) offers a non-invasive alternative, and EEG-to-text (EEG2Text) has been widely explored. Interestingly, however, EEG2Text models generally rely on teacher-forcing evaluation; without it, they fail to generate meaningful decoding. This reliance prevents EEG2Text from being applied in real-world, non-academic settings. This has fueled numerous debates about whether EEG2Text is a meaningful direction, by extension, and whether EEG truly contains decodable linguistic information. Here, using a neuropsychology-informed paradigm, we find that existing EEG2Text benchmarks have neglected EEG instability, a flaw that has confounded inference and sparked debate. Our experiments furnish key evidence for the feasibility of teacher-forcing-free EEG2Text decoding. Accordingly, we assemble the \textbf{Corpus OF Eeg-To-Text (COFETT)} using a 128-channel high-density EEG cap, providing a benchmark dedicated to evaluating EEG2Text models. In comparisons with multiple existing benchmarks, COFETT achieves SOTA ability to distinguish among model performances and enables robust, teacher-forcing-free evaluation, thereby opening a path toward practical EEG2Text applications. COFETT is open sourced in \url{https://github.com/baoyudu/COFETT}.
\end{abstract}

\section{Introduction}

Brain–computer interfaces (BCIs) that translate neural activity into text have made notable progress in restoring communication for people with severe paralysis \cite{silva2024speech}. In particular, invasive systems based on ECoG can directly decode Inner Speech, enabling near-real-time transformation of cortical signals into verbal output with high accuracy \cite{willett2023high,card2024accurate}. However, the requirement for craniotomy and implanted electrodes restricts use to patients who cannot undergo surgery and raises ethical and practical constraints.

\begin{figure}[t]
  \includegraphics[width=\linewidth]{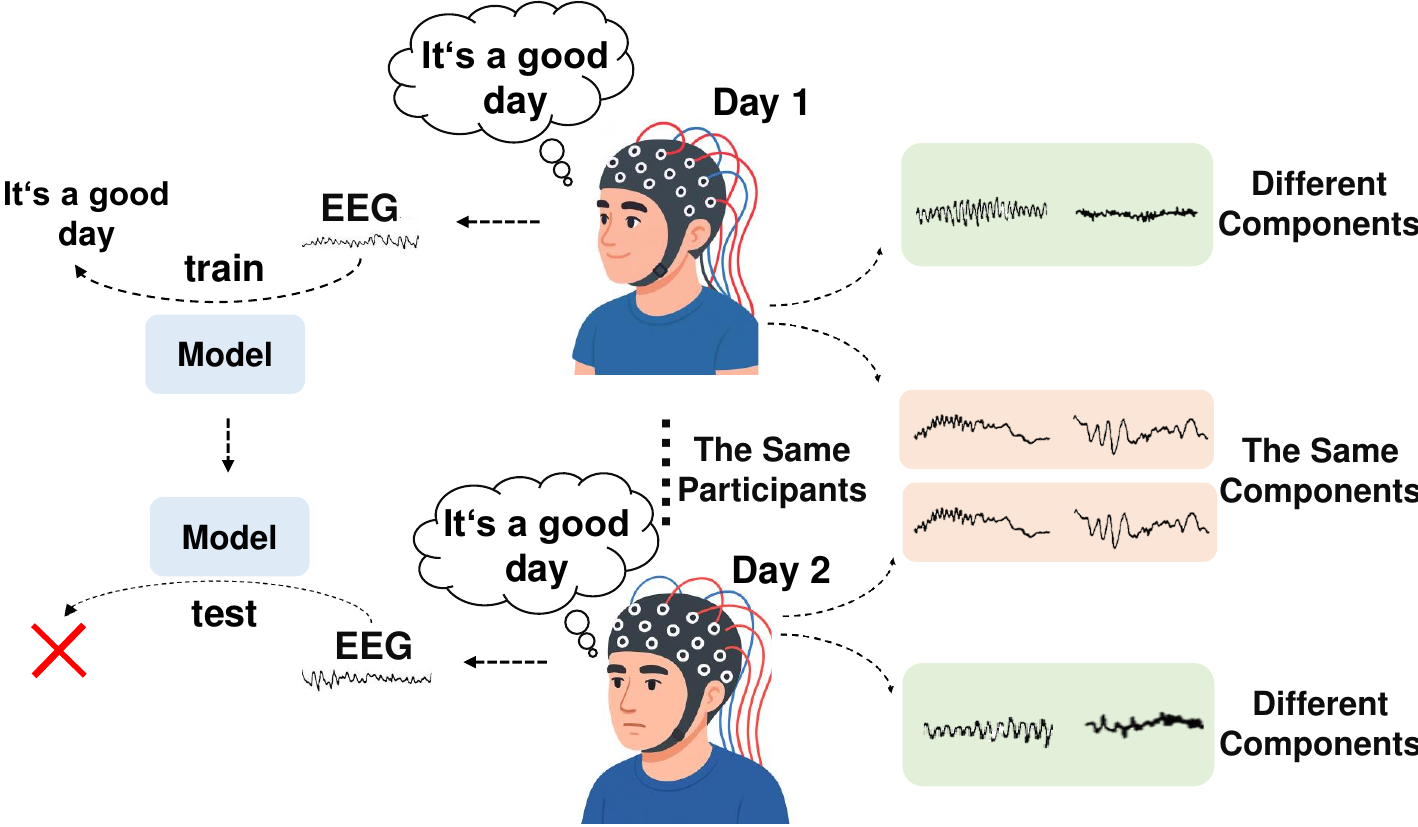}
  \caption {EEG instability in EEG2Text. The EEG elicited by imagining the same sentence on day 1 versus day 2 exhibits different components and features.}
  \label{f2}
\end{figure}

As a non-invasive alternative, several studies have attempted EEG2Text decoding, aiming to recover internal semantic intent without overt speech or motor commands \cite{wang2022open,zhou2024belt,feng2023aligning,duan2023dewave,xi2023unicorn,wang2024enhancing}. These studies have enabled EEG2Text to achieve performance comparable to ECoG2Text. So why are today's BCI applications predominantly focused on the skull-penetrating ECoG, rather than the less harmful EEG? Because these evaluations rely on teacher-forcing, a strategy appropriate for training but inappropriate for evaluation: it evaluates sequence models by feeding the ground-truth previous token to predict the next one, thereby masking exposure bias and inflating reported performance. However, it is evident that such ground-truth tokens are unavailable in real-world applications. Given this, what is the real-world performance of EEG2Text?

Recent analyses suggest that, under teacher-forced evaluation, some EEG2Text systems perform comparably when fed random noise inputs, while failing to generate meaningful decoding without teacher-forcing, casting doubt on whether models truly learn linguistically meaningful structure from EEG \cite{murad2024unveiling,jo2024eeg}. These concerns have led to broader skepticism about the feasibility of EEG2Text and underscore the need for stricter evaluation protocols.

Motivated by these challenges, we critically re-examined existing EEG2Text studies, observing that prior debates and investigations centred on model architectures while the reliability of the benchmarks themselves remained largely unexamined. We found that existing benchmarks (for example, ZuCo \cite{hollenstein2018zuco,hollenstein2019zuco}) do not consider EEG instability. EEG is inherently stochastic and markedly non-stationary even within the same participant \cite{downey2018intracortical}. Here, ‘instability’ denotes trial-to-trial and session-to-session shifts in signal statistics and spatial topography, arising from cognitive-state drift, electrode displacement, impedance changes, and physiological fluctuations. As shown in Figure \ref{f2}, a model trained on Day 1 data fails to perform effectively on data from the same subject collected on Day 2.

To address these issues, we introduce a neuropsychology-informed paradigm, which collects labelled data from multi-round inner-speech imagery with carefully spaced repetitions to maximise the recoverable linguistic component in EEG. Using 128-channel high-density recordings, we construct COFETT, a benchmark that enables teacher-forcing-free evaluation. In comparative experiments across multiple methods, COFETT shows stronger discriminative power, providing a more reliable benchmark for assessing EEG2Text model.

Our main contributions are summarised below:

\begin{itemize}
  \item \textbf{COFETT dataset.} A novel, high-density, neuropsychology-grounded EEG dataset, specifically designed for EEG2Text decoding, which accounts for EEG instability.
  \item \textbf{Teacher-forcing-free evaluation.} An assessment framework that prohibits teacher-forcing during inference, thereby measuring genuine learning.
  \item \textbf{Feasibility evidence.} Experimental results show that EEG contains linguistically decodable information, providing new evidence for the feasibility of EEG2Text decoding.
\end{itemize}

\section{Related Work}

\subsection{EEG Instability}

EEG instability, defined as temporal drift within the same participant, is a key contributor to EEG noise. It is driven by changes in attention, vigilance and fatigue, by electrode placement and impedance, and by muscle and eye movements \cite{downey2018intracortical}. As a result, recordings collected earlier and later in a dataset can follow extremely different distributions, which greatly hinders decoder training.

Because such instability is intrinsic to EEG, other BCI subfields have developed stability-aware methods and protocols, for example stationary Common Spatial Patterns that bias spatial filters toward invariant subspaces \cite{samek2012stationary}, and inter-session data-space or domain adaptation to reduce distribution shift across days \cite{arvaneh2013eeg}; both approaches improve robustness over time.

By contrast, as a nascent, deep-learning–driven field, EEG2Text relies on benchmarks such as ZuCo and ZuCo 2.0 \cite{hollenstein2018zuco,hollenstein2019zuco} that were built for natural reading and do not provide within-participant, repeated readings of the same sentences across sessions, so stability is rarely measured or exploited. This gap has slowed progress in the field.

\begin{figure*}[t]
  \centering
  \includegraphics[width=0.85\linewidth]{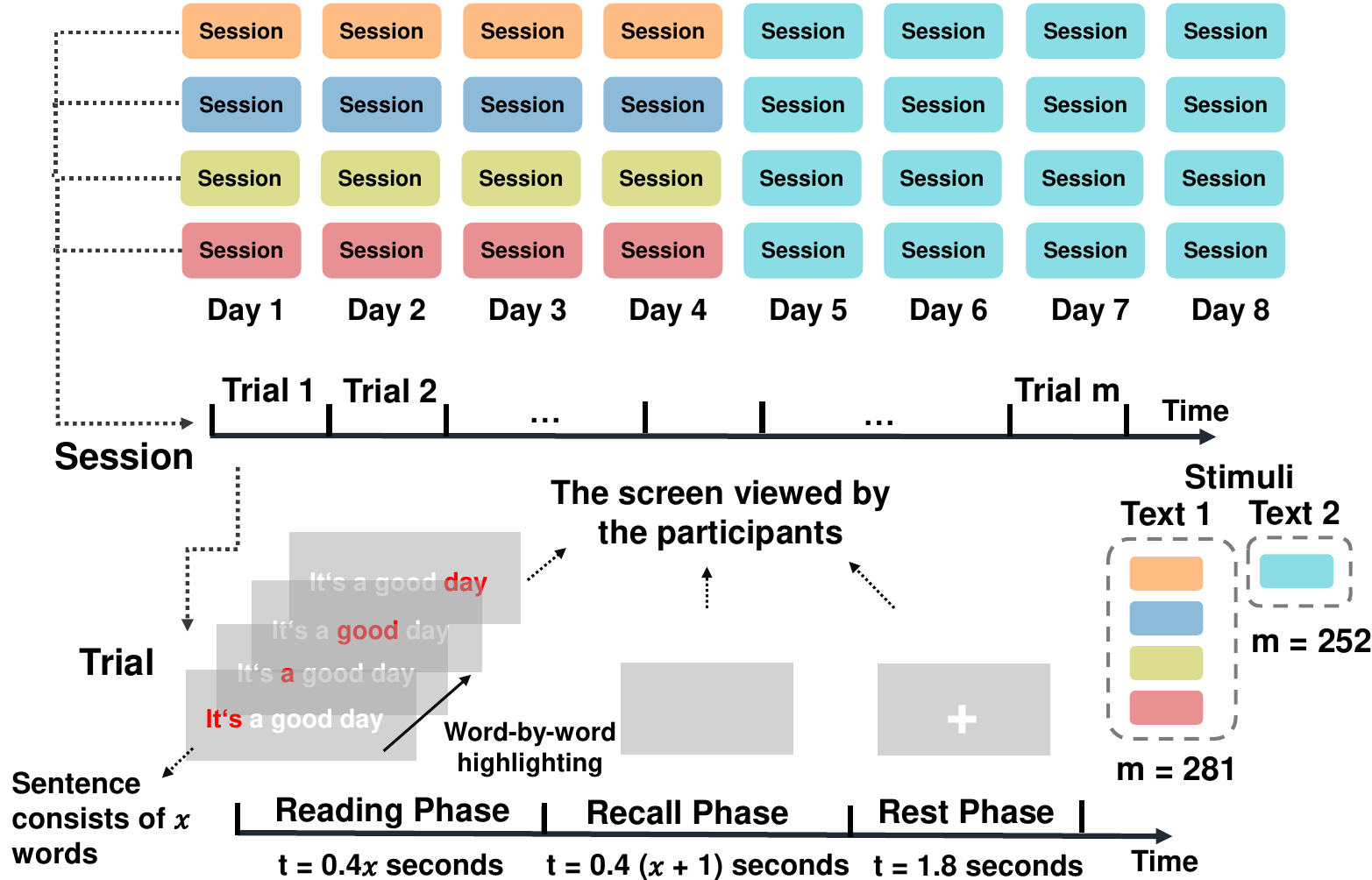}
  \caption {The full experimental procedure for each participant ($8 \text{ days} \times 4 \text{ sessions/day} \times m \text{ trials/session} \times 3 \text{ phases/trial}$
). Identical colors indicate that the participant read and imagined (recalled) the same text during that period. Specifically, days 1 to 4 involved a set of texts (Text 1), repeated 4 times, while days 5 to 8 involved a different set of texts (Text 2), repeated 16 times.}
  \label{f1}
\end{figure*}

\subsection{Teacher-Forcing EEG2Text}

Teacher-forcing feeds the ground-truth previous token to an autoregressive model when predicting the next token. During training, it stabilizes likelihood optimization and accelerates convergence for RNN/decoder-style architectures \cite{williams1989learning}. When applied at evaluation, a use that extends beyond its original purpose, it creates a train–test mismatch. Real generation must condition on the model’s own history, but conditioning on the gold history hides error accumulation, commonly referred to as exposure bias, and yields optimistically high token-level metrics.

In EEG2Text, because of the instability of EEG, several representative systems have to rely on teacher-forced evaluation; without it, EEG2Text models cannot generate useful information that can be meaningfully assessed. This practice essentially redefines generation as next-token classification conditioned on the gold context. For example, DeWave explicitly states that its evaluation uses teacher-forcing to eliminate accumulation error, turning decoding into word-level classification given the previous gold token \cite{duan2023dewave}. Subsequent work also acknowledges that reported gains still depend on teacher-forcing \cite{wang2024enhancing}. Independent analyses further show that, under teacher-forced evaluation, some pipelines perform similarly when the EEG input is replaced by random noise. This finding suggests that many decoded tokens primarily reflect the pretrained language model’s prior rather than linguistic content in EEG \cite{jo2024eeg}. Together, these observations underscore the need for stricter, teacher-forcing-free protocols.

\section{Methods}
Existing benchmarks could not evaluate model performance without teacher forcing. To enable genuine, teacher-forcing-free assessment, our benchmark makes two core advances. First, the dataset: we collect EEG from participants who repeat the same sentence at carefully spaced intervals to counter EEG instability (Methods \ref{3.1}). Second, the evaluation framework: we replace BLEU with a vector-alignment metric that evaluates teacher-forcing-free outputs without gold conditioning (Methods \ref{3.2}). Additionally, to furnish evidence for the feasibility of EEG2Text, we analyse large-scale statistical properties of EEG and conduct scaling-law analyses inspired by the LLM literature (Methods \ref{3.3}).

\subsection{Construction of COFETT}
\label{3.1}
In EEG2Text, instability has a narrower definition because we focus only on the language-related components of EEG. Figure \ref{f2} illustrates this: instability refers to the variant signal components when the same individual imagines the same sentence at different time points. Although EEG2Text is formally framed as a machine translation task, it differs fundamentally from conventional machine translation. Traditional translation involves relatively fixed and invariant mappings between sentences in two languages. By contrast, EEG2Text requires mapping highly variable neural signals to sentences in natural language. Unlike static sentence representations, EEG signals are dynamic, time-varying features that evolve with cognitive state and recording conditions. Therefore, the core challenge for EEG2Text lies in collecting EEG datasets that minimise the noise introduced by such instability.

Therefore, guided by insights from neuropsychology \cite{proix2022imagined}, we designed an experimental paradigm as shown in Figure \ref{f1}. Participants completed an eight-day experimental protocol, with each day consisting of four sessions. Each session included \(m\) trials, where \(m=281\) on days 1--4 and \(m=252\) on days 5--8. Each trial, in turn, was composed of word-by-word reading, word-by-word recall, and a rest period. The textual materials span 39 categories, covering nearly all routine expressions.

\begin{figure}[t]
  \includegraphics[width=\linewidth]{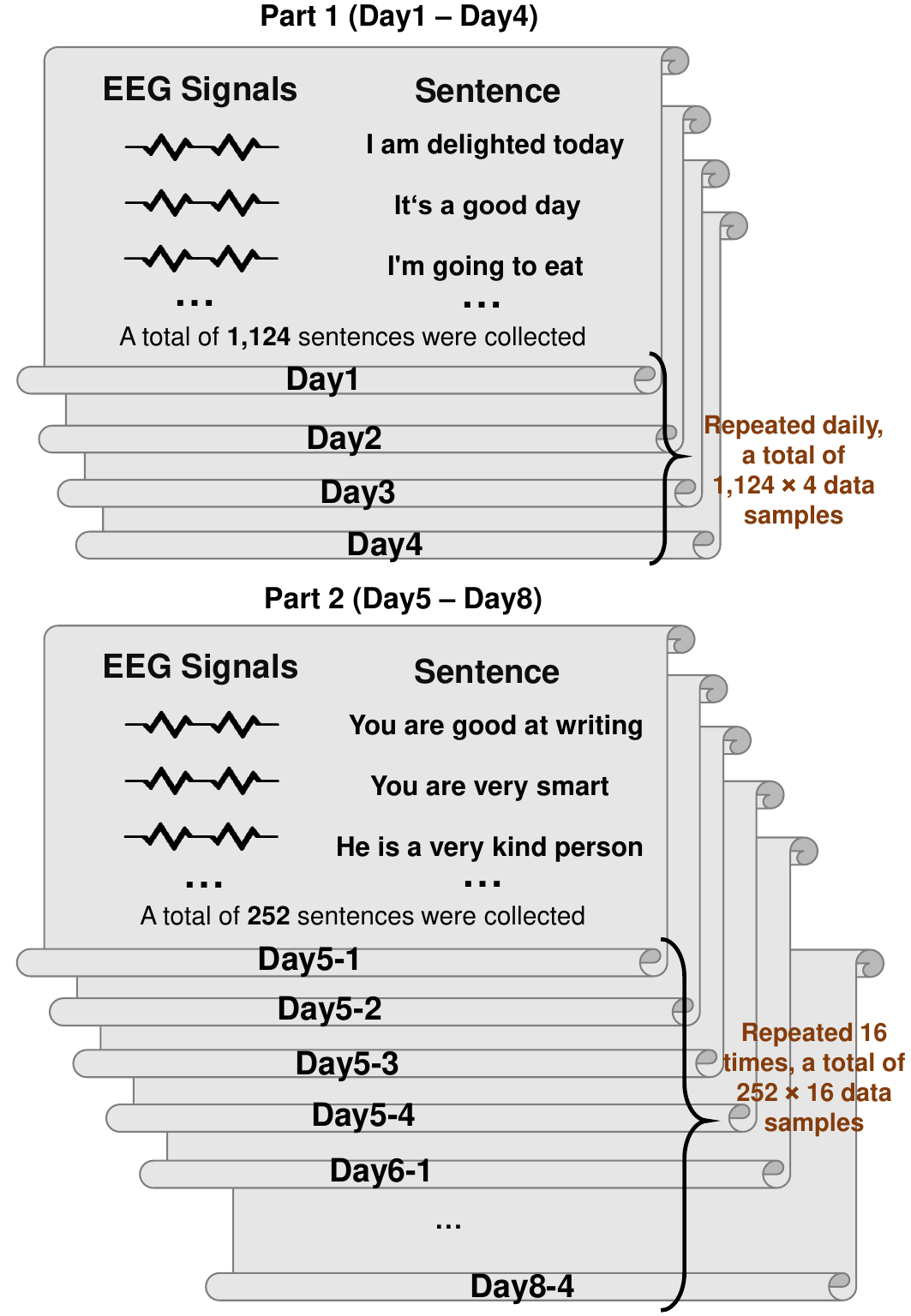}
  \caption {The EEG dataset obtained from the experiment is divided into 2 main parts. In the first part, each sentence was repeated 4 times, capturing EEG signals from the same participant recalling the identical sentence on 4 occasions. In the second part, sentences were repeated 16 times.}
  \label{f3}
\end{figure}

The word-by-word recall phase is the focus of our subsequent analysis, as it represents the Inner Speech. At this stage, participants were instructed to recall, verbatim, the content they had just read during the reading phase, with the intention of verbalizing it but refraining from doing so. The Inner Speech produced in this process aligns with the principles of EEG2Text \cite{zhang2024chisco}. 

The experimental design for single trials is inspired by previous psycholinguistic studies on imagined speech, where reading guides the imagination process. The imagined speech is considered to be embedded within the participant's Inner Speech \cite{zhang2024chisco,nieto2022thinking}. Moreover, the reading phase activates brain regions associated with language processing, such as the left temporal lobe and inferior frontal gyrus, which are also involved in Inner Speech, showing partial overlap in neural representations \cite{tian2010mental,hickok2007cortical}. Thus, the experimental paradigm in this study offers two main advantages: first, it constrains the content of participants' imagination, mitigating dataset bias that could arise from free imagination; second, the activation of similar brain regions during both phases aids in a smoother transition to the Inner Speech state. These factors collectively contribute to more effective training of EEG2Text models.

Ultimately, we obtained the EEG data shown in Figure \ref{f3}, which can also be represented as the following:
\[
\begin{aligned}
&\mathcal{D} = \{(EEG_i, S_i) \mid i = 1, 2, \dots, M, \\
&EEG_i = EEG_{i1}, EEG_{i2}, \dots, EEG_{ix}\},
\end{aligned}
\]
where \( EEG_i \) denotes a set of EEG signals associated with sentence \( S_i \), and $EEG_{i,j}$ corresponds to the EEG recording obtained from the $j$-th repetition of sentence $S_i$ by a given participant. \( S_i \) is a sentence chosen from a predefined sentence library \( \mathcal{S} = \{ S_1, S_2, \dots, S_M \} \), where each \( S_i \) is a natural language sentence.

\begin{table*}[t]
\centering
\small
\renewcommand{\arraystretch}{1.15}
\setlength{\tabcolsep}{1.4pt}
\begin{tabular}{@{}ll*{15}{c}@{}}
\toprule
& \multicolumn{15}{c}{\textbf{ZuCo (1.0+2.0)}}& \textbf{COFETT} \\
\cmidrule(lr){2-16}
\multirow{2}{*}{\textbf{Model}} & \multirow{2}{*}{\textbf{Input}} &
\multicolumn{2}{c}{\textbf{BLEU-1}} &
\multicolumn{2}{c}{\textbf{BLEU-2}} &
\multicolumn{2}{c}{\textbf{BLEU-3}} &
\multicolumn{2}{c}{\textbf{BLEU-4}} &
\multicolumn{2}{c}{\textbf{ROUGE-1 F (\%)}} &
\multicolumn{2}{c}{\textbf{WER (\%)}} &
\multicolumn{2}{c}{\textbf{Emb.(\%)}} &
\textbf{Emb.(\%)}\\
\cmidrule(lr){3-4}\cmidrule(lr){5-6}\cmidrule(lr){7-8}\cmidrule(lr){9-10}
\cmidrule(lr){11-12}\cmidrule(lr){13-14}\cmidrule(lr){15-16}
& & \textbf{w/o tf} & \textbf{w/tf} & \textbf{w/o tf} & \textbf{w/tf} &
\textbf{w/o tf} & \textbf{w/tf} & \textbf{w/o tf} & \textbf{w/tf} &
\textbf{w/o tf} & \textbf{w/tf} & \textbf{w/o tf} & \textbf{w/tf} &
\textbf{w/o tf} & \textbf{w/tf} & \textbf{w/o tf}\\
\midrule
\textbf{Brain-}
& \textbf{EEG}   & 13.75 & \underline{39.12} & 2.88 & \underline{21.98} & 0.81 & \underline{12.46} & 0.37 & \underline{7.22}& 11.84 & \underline{28.55} & 108.24 & \underline{77.97} & 5.98 & 63.89 & \textbf{14.92}\\
\textbf{BART}
& \textbf{Noise} & 14.18 & \underline{39.50} & 2.94 & \underline{22.23} & 0.88 & \underline{12.45} & 0.36 & \underline{7.17}& 11.17 & \underline{28.17} & 111.10 & \underline{78.17} & 5.96 & 64.05 & \textbf{5.96}\\
\midrule

\multirow{2}{*}{\textbf{BELT}}
& \textbf{EEG}   & 15.55 & \underline{42.04} & 4.68 & \underline{25.06} & 1.42 & \underline{13.91} & 0.46 & \underline{8.21}& 13.75 & \underline{32.53} & 110.09 & \underline{74.24} & 6.52 & 62.47 & \textbf{16.72}\\
& \textbf{Noise} & 15.52 & \underline{42.16} & 4.35 & \underline{25.03} & 1.09 & \underline{13.95} & 0.55 & \underline{8.29}& 13.83 & \underline{32.83} & 109.82 & \underline{75.38} & 6.40 & 62.28 & \textbf{6.40}\\
\midrule

\multirow{2}{*}{\textbf{Dewave}}
& \textbf{EEG}   & 14.36 & \underline{41.25} & 3.83 & \underline{24.04} & 1.13 & \underline{13.90} & 0.56 & \underline{8.22}& 12.83 & \underline{30.64} & 109.88 & \underline{79.26} & 6.01 & 63.62 & \textbf{15.21}\\
& \textbf{Noise} & 14.42 & \underline{41.62} & 3.96 & \underline{24.24} & 1.03 & \underline{14.07} & 0.38 & \underline{8.37}& 13.03 & \underline{30.87} & 110.06 & \underline{79.25} & 5.92 & 63.88 & \textbf{5.92}\\
\midrule

\multirow{2}{*}{\textbf{Pegasus}}
& \textbf{EEG}   &  8.39 & \underline{38.09} & 2.57 & \underline{21.20} & 0.89 & \underline{11.69} & 0.34 & \underline{6.07}&  0.10 & \underline{28.29} &  99.87 & \underline{78.38} & 4.52 & 61.10 & \textbf{11.76}\\
& \textbf{Noise} &  9.14 & \underline{39.12} & 2.53 & \underline{21.58} & 0.98 & \underline{11.87} & 0.14 & \underline{6.12}&  0.10 & \underline{29.12} &  99.09 & \underline{78.13} & 3.74 & 61.38 & \textbf{3.74}\\
\midrule

\multirow{2}{*}{\textbf{T5}}
& \textbf{EEG}   & 16.59 & \underline{43.33} & 5.83 & \underline{25.60} & 2.12 & \underline{15.25} & 0.75 & \underline{8.76}& 11.76 & \underline{24.96} & 111.17 & \underline{81.23} & 7.21 & 64.85 & \textbf{16.88}\\
& \textbf{Noise} & 15.50 & \underline{43.60} & 5.10 & \underline{25.60} & 1.79 & \underline{15.34} & 0.77 & \underline{8.86}& 11.01 & \underline{25.29} & 111.66 & \underline{81.48} & 6.31 & 64.32 & \textbf{6.31}\\
\midrule

\multirow{2}{*}{\textbf{EEGnet}}
& \textbf{EEG}   &  7.30 & -- & 2.05 & -- & 0.92 & -- & 0.38 & -- & 0.08 & -- & 120.06 & -- & 5.47 & -- & \textbf{16.29}\\
& \textbf{Noise} &  7.54 & -- & 2.05 & -- & 0.92 & -- & 0.38 & -- & 0.08 & -- & 121.06 & -- & 5.02 & -- & \textbf{5.02}\\
\midrule

\textbf{EEG-}
& \textbf{EEG}  & 12.54 & -- & 2.86 & -- & 0.61 & -- & 0.12 & -- & 10.11 & -- & 105.01 & -- & 1.97 & -- & \textbf{9.13}\\
\textbf{Conformer}
& \textbf{Noise} & 11.30 & -- & 2.83 & -- & 0.26 & -- & 0.27 & -- & 9.65 & -- & 106.13 & -- & 1.44 & -- & \textbf{1.44}\\
\midrule

\multirow{2}{*}{\textbf{EEGPT}}
& \textbf{EEG}   & 16.34 & \underline{43.00} & 6.19 & \underline{26.77} & 1.76 & \underline{15.29} & 0.80 & \underline{8.92}& 14.92 & \underline{31.96} & 109.85 & \underline{80.09} & 7.04 & 65.01 & \textbf{17.13}\\
& \textbf{Noise} & 16.04 & \underline{43.34} & 5.98 & \underline{26.90} & 1.81 & \underline{15.18} & 0.98 & \underline{9.04}& 14.73 & \underline{31.20} & 109.92 & \underline{80.11} & 6.51 & 64.55 & \textbf{6.51}\\

\bottomrule
\end{tabular}
\caption{Comparison of models on ZuCo and COFETT. `w/o tf' denotes evaluation without teacher-forcing; `w/tf' denotes evaluation with teacher-forcing. `Emb.' refers to the Pearson correlation between EEG embeddings and text embeddings (Section \ref{3.2}). Underlined entries denote metrics commonly used in existing work; bold entries indicate our proposed methods; the remaining entries are our extensions of existing baselines.}
\label{zuco}
\end{table*}

\subsection{Teacher-Forcing-Free Evaluation}
\label{3.2}

\begin{figure}[t]
  \includegraphics[width=\linewidth]{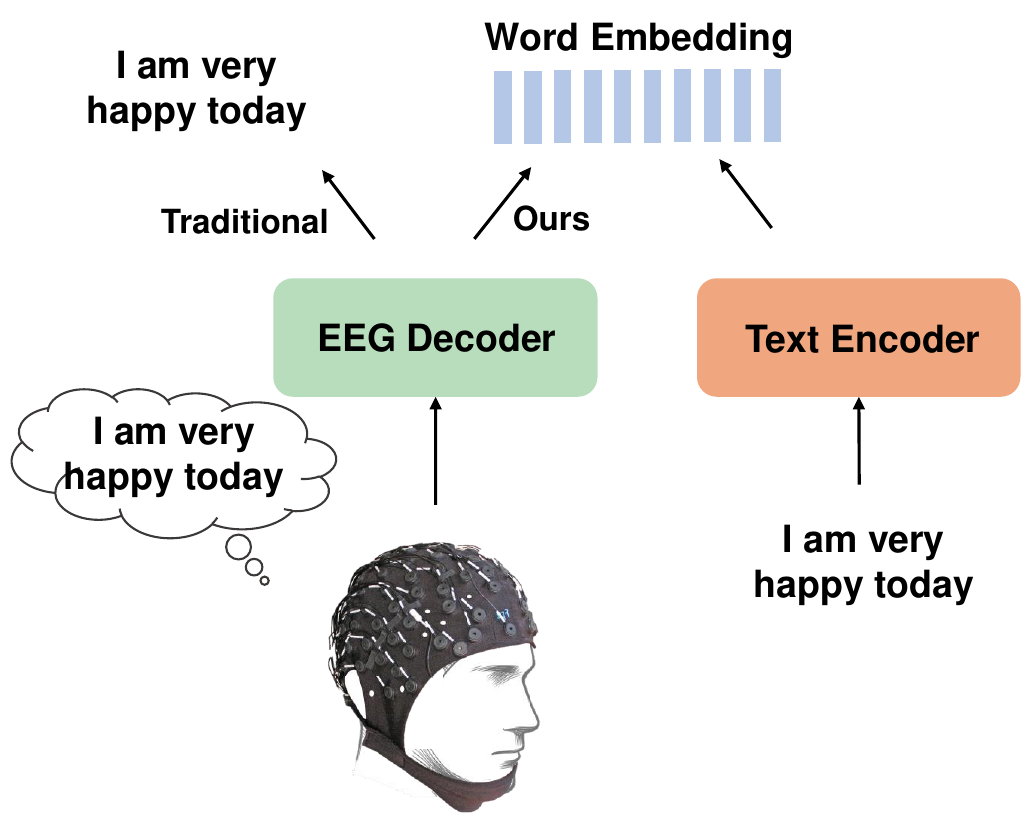}
  \caption {Comparison of supervision methods for EEG2Text models.}
  \label{f4}
\end{figure}

Existing work shows that, without teacher-forcing, end-to-end EEG2Text models fail to extract meaningful information from the training data irrespective of architecture \cite{jo2024eeg}. We therefore take an approach inspired by reinforcement learning. When the ultimate task becomes too challenging, causing the model to engage in passive learning, we can pragmatically decompose the overarching goal into smaller, more manageable sub-goals. In tasks such as machine translation or EEG2Text, the semantic distance in the embedding space can serve as a proxy for this decomposition. Semantic distance is a key metric for evaluating the performance of multilingual models. It measures the distance in embedding space between different languages under similar semantic conditions, serving as an important indicator of machine translation performance \cite{bajpai2024multilingual}. 

Accordingly, we adopt the model shown in Fig. \ref{f4}. Rather than training EEG2Text with text tokens as the supervision signal, we supervise the model using text embeddings from a pretrained language model. At test time, we evaluate performance by the Pearson correlation between EEG embeddings and text embeddings, instead of BLEU between text outputs. To prevent data leakage, EEG data in the training and test sets were drawn from different sessions of the same participant, and a separate model was trained for each participant.

\subsection{Feasibility Evidence}
\label{3.3}
We test whether EEG carries linguistic information by comparing the similarity of signals when the same participant imagines the same sentence at different times with the similarity when the same participant imagines different sentences at different times. COFETT uniquely enables this analysis by providing within-participant, repeated renditions of identical sentences—capability absent from existing benchmarks. We quantify EEG similarity using `MINDFUL' \cite{pun2024measuring} and `CCA' \cite{dmochowski2012correlated}, with full mathematical details in the Appendix.

From a complementary angle, we also apply an LLM-style scaling analysis \cite{kaplan2020scaling,henighan2020scaling,hoffmann2022training}: we vary the proportion of training data and examine whether model performance scales predictably with data size. Subsequent Experiments report the results.

\begin{table*}[t]
\centering
\small
\renewcommand{\arraystretch}{1.15}
\setlength{\tabcolsep}{1.2pt}
\begin{tabular}{@{}ll*{14}{c}@{}}
\toprule
\multirow{2}{*}{\textbf{Model}} & \multirow{2}{*}{\textbf{Input}} &
\multicolumn{2}{c}{\textbf{BLEU-1}} &
\multicolumn{2}{c}{\textbf{BLEU-2}} &
\multicolumn{2}{c}{\textbf{BLEU-3}} &
\multicolumn{2}{c}{\textbf{BLEU-4}} &
\multicolumn{2}{c}{\textbf{ROUGE-1 F (\%)}} &
\multicolumn{2}{c}{\textbf{WER (\%)}} &
\multicolumn{2}{c}{\textbf{Emb.}}\\
\cmidrule(lr){3-4}\cmidrule(lr){5-6}\cmidrule(lr){7-8}\cmidrule(lr){9-10}%
\cmidrule(lr){11-12}\cmidrule(lr){13-14}\cmidrule(lr){15-16}
& & \textit{w/o rep} & \textit{w/ rep}
  & \textit{w/o rep} & \textit{w/ rep}
  & \textit{w/o rep} & \textit{w/ rep}
  & \textit{w/o rep} & \textit{w/ rep}
  & \textit{w/o rep} & \textit{w/ rep}
  & \textit{w/o rep} & \textit{w/ rep}
  & \textit{w/o rep} & \textit{w/ rep} \\
\midrule
\textbf{Brain-}
& \textbf{EEG}   & 13.77 & 14.36 & 2.90 & 3.03 & 0.82 & 0.95 & 0.36 & 0.42 & 11.83 & 11.33 & 108.26 & 110.30 & 6.03  & \textbf{14.92}\\
\textbf{BART}
& \textbf{Noise} & 14.18 & 14.18 & 2.94 & 2.94 & 0.88 & 0.88 & 0.36 & 0.36 & 11.17 & 11.17 & 111.10 & 111.10 & 5.96  & \textbf{5.96}\\
\midrule

\multirow{2}{*}{\textbf{BELT}}
& \textbf{EEG}   & 15.57 & 15.73 & 4.65 & 4.47 & 1.44 & 1.20 & 0.47 & 0.64 & 13.77 & 14.00 & 110.07 & 109.10 & 6.49  & \textbf{16.72}\\
& \textbf{Noise} & 15.52 & 15.52 & 4.35 & 4.35 & 1.09 & 1.09 & 0.55 & 0.55 & 13.83 & 13.83 & 109.82 & 109.82 & 6.40  & \textbf{6.40}\\
\midrule

\multirow{2}{*}{\textbf{Dewave}}
& \textbf{EEG}   & 14.35 & 14.58 & 3.85 & 4.06 & 1.12 & 1.12 & 0.55 & 0.46 & 12.84 & 13.21 & 109.90 & 109.41 & 5.98  & \textbf{15.21}\\
& \textbf{Noise} & 14.42 & 14.42 & 3.96 & 3.96 & 1.03 & 1.03 & 0.38 & 0.38 & 13.03 & 13.03 & 110.06 & 110.06 & 5.92  & \textbf{5.92}\\
\midrule

\multirow{2}{*}{\textbf{Pegasus}}
& \textbf{EEG}   &  8.41 &  9.27 & 2.56 & 2.60 & 0.90 & 1.04 & 0.33 & 0.19 &  0.11 & 0.16 &  99.86 &  98.54 & 4.32  & \textbf{11.76}\\
& \textbf{Noise} &  9.14 &  9.14 & 2.53 & 2.53 & 0.98 & 0.98 & 0.14 & 0.14 &  0.10 & 0.10 &  99.09 &  99.09 & 3.74  & \textbf{3.74}\\
\midrule

\multirow{2}{*}{\textbf{T5}}
& \textbf{EEG}   & 16.58 & 15.70 & 5.85 & 5.22 & 2.13 & 1.90 & 0.76 & 0.85 & 11.78 & 11.16 & 111.19 & 110.76 & 7.08  & \textbf{16.88}\\
& \textbf{Noise} & 15.50 & 15.50 & 5.10 & 5.10 & 1.79 & 1.79 & 0.77 & 0.77 & 11.01 & 11.01 & 111.66 & 111.66 & 6.31  & \textbf{6.31}\\
\midrule

\multirow{2}{*}{\textbf{EEGnet}}
& \textbf{EEG}   &  7.32 & 7.62 & 2.05 & 2.12 & 0.94 & 0.98 & 0.36 & 0.41 & 0.10 & 0.11 & 120.05 & 119.80 & 5.67 &  \textbf{16.29}\\
& \textbf{Noise} &  7.54 & 7.54 & 2.05 & 2.05 & 0.92 & 0.92 & 0.38 & 0.38 & 0.08 & 0.08 & 121.06 & 121.06 & 5.02 &  \textbf{5.02}\\
\midrule

\textbf{EEG-}
& \textbf{EEG}  & 12.56 & 11.45 & 2.88 & 2.92 & 0.63 & 0.32 & 0.12 & 0.30 & 10.13 & 9.82 & 105.00 & 104.60 & 2.01 &  \textbf{9.13}\\
\textbf{Conformer}
& \textbf{Noise} & 11.30 & 11.30 & 2.83 & 2.83 & 0.26 & 0.26 & 0.27 & 0.27 & 9.65 & 9.65 & 106.13 & 106.13 & 1.44  & \textbf{1.44}\\
\midrule

\multirow{2}{*}{\textbf{EEGPT}}
& \textbf{EEG}   & 16.33 & 16.23 & 6.20 & 6.08 & 1.77 & 1.89 & 0.79 & 1.05 & 14.93 & 14.90 & 109.86 & 109.12 & 6.96  & \textbf{17.13}\\
& \textbf{Noise} & 16.04 & 16.04 & 5.98 & 5.98 & 1.81 & 1.81 & 0.98 & 0.98 & 14.73 & 14.73 & 109.92 & 109.92 & 6.51  & \textbf{6.51}\\
\bottomrule
\end{tabular}
\caption{COFETT Benchmark ablation results. w/o rep denotes the ablation group without within-participant repetitions, whereas w/ rep denotes the control group with within-participant repetitions.}
\label{ablation}
\end{table*}

\section{Experiments}
\subsection{Benchmark Comparison}

Most existing EEG2Text studies evaluate on the ZuCo corpus and report BLEU, ROUGE-1 or WER under teacher-forcing. Accordingly, we conducted controlled comparisons on ZuCo 1.0/2.0, replacing EEG with pure random noise at both training and test time to assess whether scores change. For models, we evaluate three EEG2Text systems, BrainBART \cite{wang2022open}, BELT \cite{zhou2024belt} and DeWave \cite{duan2023dewave}, two general-purpose Transformer baselines, Pegasus \cite{zhang2020pegasus} and T5 \cite{raffel2020exploring}, two EEG feature-extraction architectures, EEGConformer \cite{song2022eeg} and EEGNet \cite{lawhern2018eegnet}, and a decoder-style large EEG foundation models, EEGPT \cite{wang2024eegpt}, to cover the full range of model classes. To extract language embeddings, we employ LaBSE \cite{feng-etal-2022-language}, a multilingual model designed for sentence-level representation.

Additionally, to strengthen the comparison, we extended the conventional setups with vector-alignment variants and non-teacher-forcing evaluation. As shown in Table \ref{zuco}, across all ZuCo 1.0/2.0 conditions, scores obtained with true EEG are not statistically distinguishable from those obtained with noise. This indicates that the canonical ZuCo benchmark primarily reflects models’ language priors, rather than their use of EEG, and thus cannot demonstrate learning of meaningful neural information.

By contrast, on our COFETT benchmark with teacher-forcing-free evaluation, performance separates clearly across methods and collapses under the same noise controls, indicating that COFETT discriminates models by their ability to exploit EEG rather than by linguistic priors.

\begin{table*}\small
\centering
\begin{tabular}{cccccccccc}
\toprule
\multirow{2}{*}{\textbf{ID}} & \multicolumn{3}{c}{\textbf{Condition}} & \multirow{2}{*}{\textbf{MINDFUL}} & \multicolumn{4}{c}{\textbf{CCA}}\\
\cmidrule(lr){2-4} \cmidrule(lr){6-9}
 & Subject & Time & Text &  & \MakeUppercase{\romannumeral 1}(\%) & \MakeUppercase{\romannumeral 2}(\%) & \MakeUppercase{\romannumeral 3}(\%) & Mean  \\
\midrule
1 & Same &        Different Day &         Same &           1.64 &     17.42 &     18.58 &     18.81 & 18.27 \\
2 & Same &        Different Day &         Different &           1.67 &     16.59 &     18.00 &     18.52 & 17.70 \\
3 & Same &       Close Time &         Same &           1.54 &     23.23 &     20.81 &     21.17 & 21.74 \\
4 & Same &       Close Time &         Different &           1.79 &     23.17 &     19.97 &     21.30 & 21.48 \\
5 & Different &     -       &         Same &           1.96 &     13.55 &     14.41 &     14.90 & 14.29 \\
6 & Different &      -      &         Different &           2.07 &     13.48 &     14.98 &     15.82 & 14.76 \\
\bottomrule
\end{tabular}
\caption{Similarity of EEG signals under various conditions. For instance, ID 1 represents the EEG similarity of the same participant attempting to articulate the same text on different days. Higher CCA values and lower MINDFUL scores indicate greater similarity between EEG signals.}
\label{similar}
\end{table*}

\subsection{Benchmark Ablations}

We base the benchmark on two design choices that jointly enable teacher-forcing-free evaluation. First, within-participant exact repetitions expose the invariant, language-bearing component of EEG by implicitly averaging over time-varying noise, thereby mitigating EEG instability and improving the statistical reliability of comparisons across methods. Second, a sentence-level embedding-alignment metric evaluates semantic compatibility between EEG-derived representations and text under teacher-forcing-free decoding, which curbs negative learning behaviors that arise with token-level or n-gram scoring and discourages solutions that exploit language priors while ignoring EEG. To verify the necessity of these ingredients, we conduct two ablations under the same teacher-forcing-free protocol. 

First, we remove within-participant repetitions and add new, unique sentences to keep the total EEG duration constant; this choice diminishes the benchmark’s discriminative power and destabilizes method rankings. Second, we replace the embedding-alignment objective with teacher-forcing-free BLEU, ROUGE, and WER, which increases dependence on the pretrained language backbone and weakens sensitivity to EEG versus noise controls. The experiments show that a clean separation between models is achieved only when both core components are retained, as summarized in Table \ref{ablation}.

\begin{figure*}[t]
  \centering
  \includegraphics[width=\linewidth]{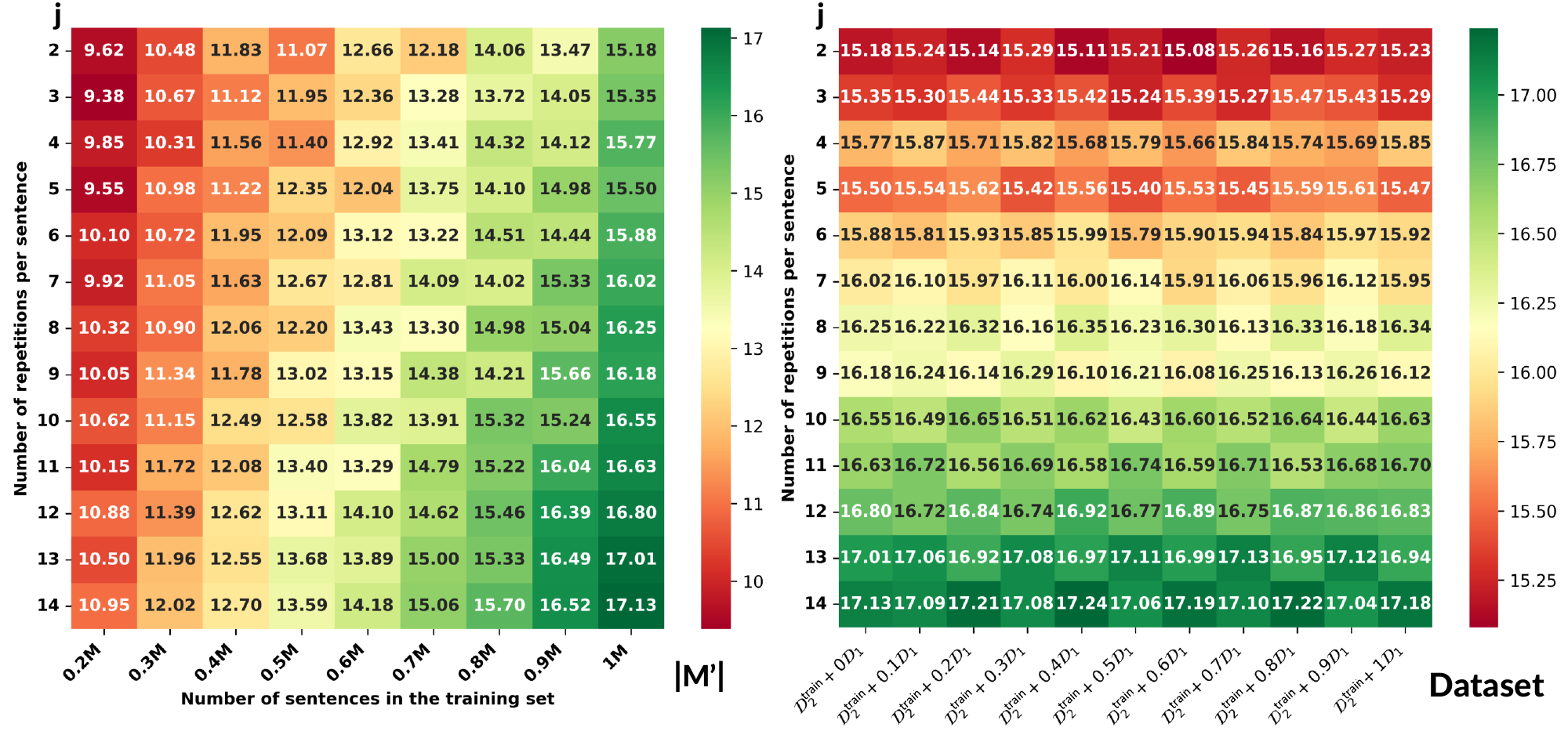}
  \caption {Alignment performance at different data scales.}
  \label{f5}
\end{figure*}

\begin{figure}[t]
  \includegraphics[width=0.85\linewidth]{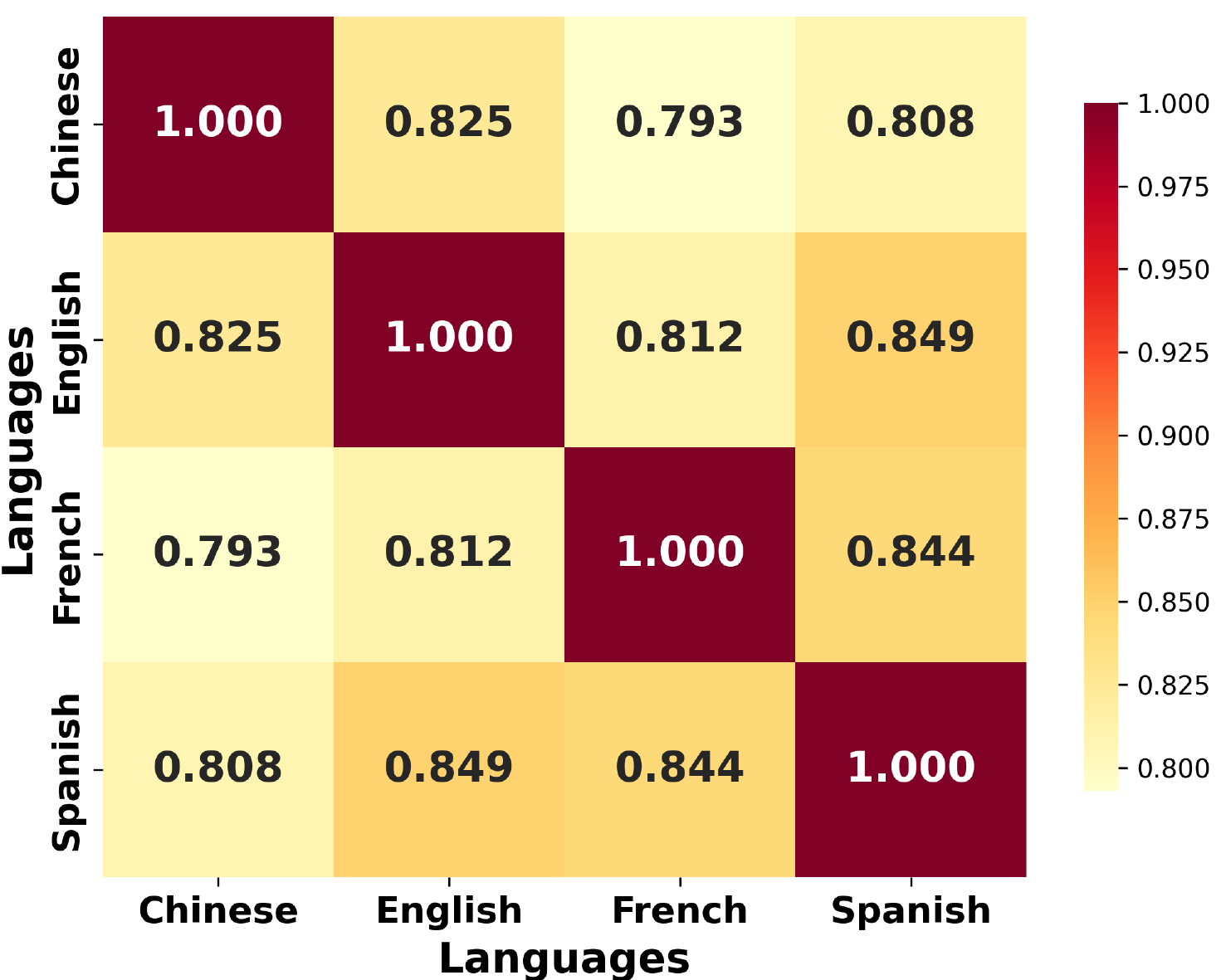}
  \centering
  \caption {Embedding distances of the same sentence expressed in different languages.}
  \label{f6}
\end{figure}

\subsection{Statistical Evidence of Linguistic Information in EEG}

To quantify whether EEG carries recoverable linguistic signal, we estimate EEG similarity with two complementary, label-free metrics and aggregate their outcomes via Monte Carlo sampling. Specifically, for many randomly drawn pairs of EEG we compute (i) MINDFUL \cite{pun2024measuring}, a recent instability/similarity measure designed for long-term neural interfaces that detects distributional changes without intention labels, and (ii) CCA \cite{dmochowski2012correlated}, which extracts components that are maximally correlated across paired recordings and thereby indexes reliable shared structure. We repeat this procedure across six pairing conditions (spanning participant identity and time separation, and controlling the sentence identity) and summarize the results (Table \ref{similar}). 

The pattern is consistent across both metrics: similarity is lowest across different participants, increases within the same participant across days, and is highest within the same participant on the same day. After conditioning on participant and time, sentence identity contributes only a small additional effect, indicating that—within our task—the time-varying physiological and hardware factors dominate the variance, whereas the language-related component is comparatively weak. This is compatible with biological accounts of organismal regulation and homeostasis \cite{bechtel2024situating}: neural systems actively resist rapid internal fluctuations, so adjacent measurements can appear similar even when the imagined sentence differs.

Taken together, these results indicate that EEG contains linguistically decodable structure, even if weak, and that decoding is feasible.

\subsection{Scaling-Law Perspective on Linguistic Information in EEG}
To further substantiate the presence of linguistic components in EEG and to quantify how within-participant repetitions and data scale affect vector alignment, we partition the corpus by recording day. For each participant, data from Days~1--4 constitute $\mathcal{D}_1$, and data from Days~5--8 constitute $\mathcal{D}_2$. We then split $\mathcal{D}_2$ into a training set $\mathcal{D}_2^{\mathrm{train}}$ and a test set $\mathcal{D}_2^{\mathrm{test}}$.
\[
\begin{aligned}
&\mathcal{D}_2^{\mathrm{train}} = \{(EEG_i, S_i) \mid i = 1, 2, \dots, M', \\
&EEG_i = EEG_{i1}, EEG_{i2}, \dots, EEG_{ij}\}.
\end{aligned}
\]
\[
\begin{aligned}
&\mathcal{D}_2^{\mathrm{test}} = \{(EEG_i, S_i) \mid i = 1, 2, \dots, M, \\
&EEG_i = EEG_{i15}, EEG_{i16}\}.
\end{aligned}
\]

Here, the test set $\mathcal{D}_2^{\mathrm{test}}$ consists of two repetitions for every sentence in the corpus, while the training set $\mathcal{D}_2^{\mathrm{train}}$ contains $j$ repetitions drawn from a subset of sentences $M'$, where $M'\subset M$. We scale data by jointly varying the subset size $|M'|$ and the repetition count $j$ while keeping $\mathcal{D}_2^{\mathrm{test}}$ fixed, enabling an LLM-style scaling-law analysis on EEG.

In this section, we select EEGPT\cite{wang2024eegpt}, the best-performing model in the previous experiments, as the EEG encoder, together with LaBSE\cite{feng-etal-2022-language} for text representation.

As shown in Figure~\ref{f5}(a), alignment performance increases with both the repetition count $j$ and the number of sentences $M'$, with the effect of $M'$ exceeding that of $j$. To examine the regime $M'>M$, we proportionally augment the training set with samples from $\mathcal{D}_1$. The results in Figure~\ref{f5}(b) show no further gains once $M'$ exceeds $M$. We attribute this to an in-/out-of-domain effect: $\mathcal{D}_2^{\mathrm{test}}$ and $\mathcal{D}_2^{\mathrm{train}}$ are in-domain when both are drawn from the same sentence set $M$; sentences added beyond $M$ are out of domain and do not improve EEG2Text performance. 

Taken together, these findings not only support the presence of linguistically decodable information in EEG, but also indicate that training and test sets must be drawn from the same semantic domain. This helps explain prior failures on ZuCo: ZuCo is a natural-reading corpus in which stories progress semantically rather than repeat semantically similar content, rendering many samples effectively out of domain with respect to one another.

\subsection{Predictive Analysis}

In this section, we explore an intriguing question: to what extent are EEG2Text and current machine translation systems comparable in performance? To examine this, we compare the embedding distances of the same sentence expressed in different languages. As shown in comparison to Figure \ref{f6}, it is clear that in existing multilingual pretraining models the semantic distance between EEG and natural language remains substantially larger than the distances observed between different natural languages.

We now provide a rough projection under the linear trend inferred above. 
If the in-domain upper envelope is $\approx 0.17$, then gains must primarily come from increasing within-participant repetitions. 
Extrapolating to $\approx 0.85$ at the lower right of Figure \ref{f6} suggests $\sim 700$ repetitions per sentence, i.e., a $\sim 40\times$ increase over our current setting. This means that EEG2Text requires $\sim 40\times$ more paired samples to approach machine-translation-like performance, without accounting for diminishing returns at larger scales and the instability of EEG over longer time scales. This estimate highlights the need for more EEG--text resources.

\section{Conclusion}

We provide a systematic synthesis of the emerging EEG2Text literature and identify a core limitation of current benchmarks: their design obscures feasibility by ignoring EEG instability, thereby fuelling the field’s central debate. In response, we introduce a new benchmark COFETT, comprising two components: a within-subject repetition paradigm and a vector-alignment evaluation protocol. These enables teacher-forcing-free assessment. Using this framework, we obtain convergent evidence that EEG2Text is a feasible and scientifically meaningful direction and the EEG2Text model holds promise for practical applications.

\section*{Limitations}

This work centres on benchmarking rather than proposing new architectures; as such, we did not explore the model-design space exhaustively. It remains possible that alternative architectures or training regimes could further reduce the data requirements observed in our predictive analyses. Nevertheless, any ultimate ceiling on performance is constrained by the signal-to-noise ratio of EEG, which imposes a theoretical upper bound on decodable linguistic information.

\section*{Ethics Statement}

This study involved human participants in the collection of EEG data. All participants provided written informed consent for participation and data reuse prior to the experiments. The study was reviewed and approved by the Ethics Commission of Harbin Institute of Technology (Approval Number: HIT2024035) and was conducted in accordance with the Declaration of Helsinki (2013). The study involved only non-invasive EEG recordings and standard cognitive tasks, and therefore posed minimal risk to participants.

\section*{Acknowledgments}

The research in this article is supported by the New Generation Artificial Intelligence of China (2024YFE0203700), National Natural Science Foundation of China under Grants U22B2059 and 62576124.

\bibliography{custom}
\newpage
\appendix

\section{Participants and pre-experiment}

\begin{table*}[h]
    \centering
    \begin{tabular}{lllllllll}
    \toprule
        \textbf{Variable} & \textbf{S1} & \textbf{S2} & \textbf{S3} & \textbf{S4} & \textbf{S5} & \textbf{S6} & \textbf{S7} \\
    \midrule
        \textbf{Gender} & male & female & male & female & male & male & female \\
        \textbf{Age} & 26 & 25 & 22 & 47 & 40 & 42 & 28 \\
        \textbf{ANT(Alerting)} & 25.1 & 20.3 & 30.5 & 55.2 & 68.9 & 50.4 & 30.5 \\
        \textbf{ANT(Orienting)} & 40.8 & 45.7 & 30.6 & 110.3 & 25.0 & 85.2 & 60.4 \\
        \textbf{ANT(Executive Control)} & 70.4 & 87.3 & 95.7 & 115.1 & 140.9 & 100.8 & 120.6 \\
        \textbf{ANT(Mean)} & 45.4 & 51.1 & 52.3 & 93.5 & 78.3 & 78.8 & 70.5 \\
        \textbf{HSK(Reading Section)} & 97.0 & 95.0 & - & - & - & - & - \\
    \bottomrule
    \end{tabular}
    \caption{Details of all subjects in the study.}
    \label{sub}
\end{table*}

EEG data were collected from two subjects: S1 (male, age 26) and S2 (female, age 25). Neither participant reported a history of neurological or psychiatric conditions. All participants were right-handed, had normal hearing, and either had normal or corrected-to-normal vision. None of them had previous experience with brain-computer interfaces (BCIs).

To improve the quality of Inner Speech EEG data, we first conducted a screening process. Seven volunteers (aged 22–47, Chinese native speakers, 4 males) were initially recruited for a five-day session of Integrative Body-Mind Training (IBMT) \cite{tang2007short}, a meditation technique designed to enhance concentration. Previous research has shown that IBMT significantly improves attention capacity and reduces fatigue. Additionally, IBMT training has been associated with increased vigor, as measured by the Profile of Mood States scale, and a marked reduction in stress-related cortisol levels \cite{tang2007short}. This training was aimed at preparing participants for the cognitively demanding experimental paradigm.

After completing the IBMT session, all participants took the Attention Network Test (ANT) \cite{fan2002testing}, a psychological assessment that evaluates attentional functions. The ANT measures three key indices: alerting (readiness for an impending event), orienting (ability to focus attention on a specific spatial location), and executive control (capacity to handle conflicting information). Lower ANT scores indicate greater attentional focus. We calculated the mean scores of all participants across the three ANT subtests and selected the two participants with the lowest scores for further experimentation. Subsequently, the two selected participants underwent a Chinese language proficiency test (HSK Level 6 Reading Section) to ensure they could accurately comprehend the text presented during the experiments. Details of each participant are provided in Table \ref{sub}. No participants withdrew during the experiment.

Upon completion of the experiment, all seven participants received a base compensation of \$50. Additionally, two participants selected to complete the full experiment received an extra \$400, which, when calculated on an hourly basis, is slightly above the average hourly wage in the region where the experiment was conducted.

\section{Textual stimuli}

The textual stimuli presented to the participants were sourced from the Chisco \cite{zhang2024chisco}, which comprises daily language texts in Chinese across 39 semantic categories. As shown in Figure \ref{f1}, our experimental paradigm consists of two sets of text data. Text 1 contains the complete set of 8 semantic categories. Text 2 includes the complete set of 2 semantic categories. For convenience, a description of the Chisco text data \cite{zhang2024chisco} has been provided as follows.

To ensure the dataset is suitable for training BCIs for everyday use, the textual materials were designed to encompass a broad range of daily language. To achieve this, Chisco manually selected expressions from the Chinese social media platform Weibo, as well as public datasets ROCstory \cite{mostafazadeh2016corpus} and Dailydialog \cite{li2017dailydialog}. These expressions were initially categorized into 39 categories using a combination of machine learning clustering algorithms and manual annotation by human experts. The expressions were then rephrased into sentences of 6 to 15 Chinese characters through a crowdsourcing approach, resulting in a dataset representing daily Chinese expressions.

The crowdsourcing process adhered to the following criteria to ensure the quality and usability of the text data:

\begin{itemize}[noitemsep]
    \item Each sentence must be independently understandable, without requiring multiple segments to form a coherent discourse.
    \item Sentences must avoid clauses and indirect anaphora.
    \item The number of nouns in a sentence must not exceed three.
    \item Ambiguous sentences were excluded.
    \item Texts containing any form of racial or gender discrimination were prohibited.
\end{itemize}

This approach was designed to minimize syntactic complexity, thereby reducing the likelihood of errors during the imagined speech process. Examples of sentences with their categories include:

\begin{itemize}[noitemsep]
    \item ``Today's dinner tasted great'' - \textit{Food and Dining}
    \item ``She is going to listen to the concert'' - \textit{Performing Arts}
    \item ``I am deeply sorry for my behavior'' - \textit{Apologies}
\end{itemize}

\section{EEG data acquisition}

The study was conducted in an electrically shielded room. Participants were seated comfortably in front of a computer screen positioned 80 cm away, directly facing its center. Stimuli were presented on a 26-inch screen with a refresh rate of 60 Hz and a resolution of $1920 \times 1080$ pixels. The visual stimuli consisted of sentences from everyday language, ranging from 6 to 15 Chinese characters in length. These sentences were displayed in a 28-point Song typeface, on a single line, centrally placed on a grey background with white text to minimize glare and eye strain. During breaks, participants were provided with snacks and water, and encouraged to rest.

To ensure consistent data quality, a personalized head case was designed for each participant with three main objectives: (1) to control head movements and reduce noise from electrode displacement; (2) to provide head support, alleviating physical fatigue during prolonged sessions; and (3) to maintain consistent electrode positioning across multiple sessions, as participants underwent 8 days of experiments. The head case facilitated the reapplication of the EEG cap, minimizing internal variance in the data.

EEG data were collected using the SynAmps-2 128-channel amplifier and the 128-channel Quik-Cap, both manufactured by Compumedics Neuroscan. The Curry 9 software, also from Compumedics Neuroscan, was used for hardware control and impedance checks before each experiment, ensuring impedances remained below 25 k$\Omega$. The conductive GREENTEK\textsuperscript{\textregistered} GT5 Gel was used to fill the gap between the scalp and the electrodes.

After each experimental session, PC2 generated and stored an EDF file containing continuous recordings from 125 EEG channels, 6 external channels (for vertical and horizontal electrooculography signals, as well as voltages at the mastoids on both sides), and marker signals. The experiments were conducted with a sampling rate of 1 kHz in alternating current (AC) mode, with an accuracy of 3 nV/LSB. Stimulus presentation was managed using the open-source software package PsychoPy \cite{peirce2019psychopy2} (v2023.2.3).

\section{Quality control of EEG data}

To help participants better adapt to the experimental paradigm, each participant underwent a pre-experiment training session prior to the main experiment. This session allowed participants to become familiar with the procedure, which involved word-by-word reading, followed by imagination, and then a rest period.

In addition, to ensure the quality of the data collected, participants' attention was monitored throughout the experiment. Random attention checks were incorporated into the paradigm: at the end of 10 randomly selected trials in each session, participants were required to verbally recall the content they had memorized. The accuracy of these recitals was evaluated by the experimenters, and a recall error was defined as a discrepancy of more than four characters from the original text. If a participant made two or more errors in these 10 checks, they were considered to have experienced lapses in concentration during that period. A rest period was then provided to help restore attention before restarting the session.

\section{EEG data preprocessing}
\label{Data preprocessing}

Preprocessing was performed in Python, primarily utilizing the MNE library \cite{gramfort2014mne}. Minimal preprocessing was applied to retain the maximum amount of valid information, allowing for flexibility in further processing tailored to specific research needs. The detailed steps of the preprocessing pipeline are described as follows, with an equivalent flowchart provided in the supplementary information.

\textbf{Resampling}: The raw data stored in the .edf file contained both EEG signals and event markers, sampled at 1,000 Hz. We resampled the data to 500 Hz. Given ongoing debates regarding the frequency bands associated with Inner Speech \cite{lee2020neural, proix2022imagined, crone2006high}, we employed a conservative downsampling approach. According to the Nyquist sampling theorem \cite{nyquist1928telegraph}, the 500 Hz sampling rate captures EEG information up to 250 Hz, which includes the ultra-high gamma band \cite{crone2006high}.

\textbf{The PREP pipeline} \cite{bigdely2015prep}: The PREP pipeline is a standardized early-stage EEG processing procedure, applied after data resampling. The key steps of the PREP pipeline are as follows:
\begin{enumerate}[noitemsep]
    \item Removing line noise without committing to a specific filtering strategy.
    \item Robust referencing of the signal relative to an estimate of the ``true'' average reference.
    \item Detection and interpolation of bad channels relative to this reference.
    \item Retaining sufficient information to allow users to re-reference or undo interpolation of specific channels.
\end{enumerate}

\textbf{Filtering}: In this step, we applied a 50 Hz notch filter to remove power-line noise and a zero-phase high-pass finite impulse response (FIR) filter with a 1 Hz cutoff. No low-pass filtering was performed, as explained in the `Resampling' section.

\textbf{Autoreject} \cite{jas2017autoreject}: The Autoreject algorithm was used to automatically identify and reject bad data segments and artifacts. This method optimizes rejection thresholds on a per-channel basis, ensuring an adaptive cleaning process that minimizes the loss of valid data.

\textbf{Remove Ocular Artifacts}: To effectively remove ocular artifacts from the EEG signals, we employed a multivariate linear regression model. In this approach, the horizontal electrooculography (H-EOG) and vertical electrooculography (V-EOG) signals were used as regressors. Specifically, by inputting the H-EOG and V-EOG signals into the regression model, we quantified the contribution of these two ocular movements to the EEG signal. The multivariate linear regression model was used to fit the linear relationship between the EOG signals and the EEG data, allowing for the estimation of their respective regression coefficients. These coefficients were then used to predict the ocular artifact components, which were subsequently subtracted from the original EEG data, resulting in purified signals. This method not only improved the signal-to-noise ratio of the EEG data but also enhanced the accuracy and reliability of subsequent neuroelectric analyses.

\textbf{Independent Component Analysis (ICA)} \cite{hyvarinen1997independent}: ICA is a widely used blind source separation method to remove artifacts from EEG signals. For our dataset, ICA was applied only to the EEG channels using the MNE implementation of extended Infomax ICA. Due to the wide frequency range of the data, we set the number of independent components to 30, which is higher than in other neural language decoding studies \cite{mou2024chineseeeg}, ensuring that the components capture the majority of relevant information. Noise component identification was conducted using MNE-ICALabel automatic annotation, followed by manual evaluation to ensure accuracy without introducing excessive manual processing.

\textbf{Data Segmentation}: The continuous EEG data were segmented into three types of segments based on the experimental paradigm: reading, Inner Speech, and meditation segments. 

\section{Mindful}
\label{Mindful}

The Mindful algorithm follows a pipeline process, beginning with feature extraction from two EEG signals, followed by dimensionality reduction using Principal Component Analysis (PCA) and Z-score normalization. The Kullback-Leibler (KL) divergence between the two segments is then computed, which can be mathematically represented as:
\[
\begin{aligned}
    &F_i = \text{Normal(PCA}( \text{Extract\_Features}(x_i))), i = 1, 2 \\
    &\mu_i = \frac{1}{N} \sum_{j=1}^{N} F_i^{(j)}\\
    &\Sigma_i = \frac{1}{N-1} \sum_{j=1}^{N} \left( F_i^{(j)} - \mu_i \right) \left( F_i^{(j)} - \mu_i \right)^\top \\
    &\text{Mindful}(x_1,x_2) = D_{\text{KL}}\left(\mathcal{N}(\mu_1, \Sigma_1) \,\|\, \mathcal{N}(\mu_2, \Sigma_2)\right)
\end{aligned}
\]
\section{Correlated Components Analysis (CCA)}
The computation of CCA involves the following six steps. The two input EEG signal sets are denoted as \( X_1 \) and \( X_2 \). The outputs from Step 6, \( \rho_1 \), \( \rho_2 \), and \( \rho_3 \), correspond to CCA-\MakeUppercase{\romannumeral 1}, CCA-\MakeUppercase{\romannumeral 2}, and CCA-\MakeUppercase{\romannumeral 3} as mentioned in Figure \ref{f2}(c).

\begin{enumerate}
    \item \textbf{Covariance Matrix Computation}
    
    \textbf{Description:}
    Compute the covariance matrices of the input data matrices \( X1 \) and \( X2 \).
    
    \textbf{Mathematical Representation:}
    Let \( T \) be the number of time steps or samples. The covariance matrices are calculated as:
    \[
    R_{11} = \frac{1}{T} X1 \cdot X1^T
    \]
    \[
    R_{12} = \frac{1}{T} X1 \cdot X2^T
    \]
    \[
    R_{22} = \frac{1}{T} X2 \cdot X2^T
    \]
    
    \item \textbf{Eigenvalue Problem Solving}
    
    \textbf{Description:}
    Construct matrix \( M \) and find its eigenvalues and eigenvectors.
    
    \textbf{Mathematical Representation:}
    \[
    M = (R_{11} + R_{22})^{-1} (R_{12} + R_{12}^T)
    \]
    Solve the eigenvalue equation:
    \[
    M \mathbf{v}_i = \lambda_i \mathbf{v}_i
    \]
    where \( \lambda_i \) are the eigenvalues and \( \mathbf{v}_i \) are the corresponding eigenvectors.
    
    \item \textbf{Extract and Sort Eigenvalues and Eigenvectors}
    
    \textbf{Description:}
    Sort the eigenvalues in descending order and reorder the eigenvectors accordingly to prioritize the most significant components.
    
    \textbf{Mathematical Representation:}
    Let \( \lambda_1 \geq \lambda_2 \geq \ldots \geq \lambda_K \) be the sorted eigenvalues, and \( \mathbf{v}_1, \mathbf{v}_2, \ldots, \mathbf{v}_K \) be the corresponding eigenvectors.
    
    \item \textbf{Extract Principal Components}
    
    \textbf{Description:}
    Select the top three eigenvectors corresponding to the largest eigenvalues to form the projection matrix \( W \).
    
    \textbf{Mathematical Representation:}
    \[
    W = [\mathbf{v}_1, \mathbf{v}_2, \mathbf{v}_3]
    \]
    
    \item \textbf{Projection into New Feature Space}
    
    \textbf{Description:}
    Project the original data matrices \( X1 \) and \( X2 \) into the new feature space defined by the projection matrix \( W \).
    
    \textbf{Mathematical Representation:}
    \[
    C1_{\text{full}} = W^T X1
    \]
    \[
    C2_{\text{full}} = W^T X2
    \]
    Here, \( C1_{\text{full}} \) and \( C2_{\text{full}} \) represent the projected data in the new feature space.
    
    \item \textbf{Compute Pearson Correlation Coefficients}
    
    \textbf{Description:}
    For each of the first three projected components, calculate the Pearson correlation coefficient between \( C1_{\text{full}} \) and \( C2_{\text{full}} \). If the standard deviation of any component is zero (indicating a constant), set the correlation coefficient to zero.
    
    \textbf{Mathematical Representation:}
    For each component \( j = 1, 2, 3 \):
    \[
    \rho_j = \text{corr}\left( C1_{\text{full}}[j, :], C2_{\text{full}}[j, :] \right)
    \]
    If \( \sigma\left( C1_{\text{full}}[j, :] \right) = 0 \) or \( \sigma\left( C2_{\text{full}}[j, :] \right) = 0 \), then set \( \rho_j = 0 \).
\end{enumerate}

\section{EEG Experimental Setup}

\begin{figure*}[t]
  \includegraphics[width=\linewidth, page=1]{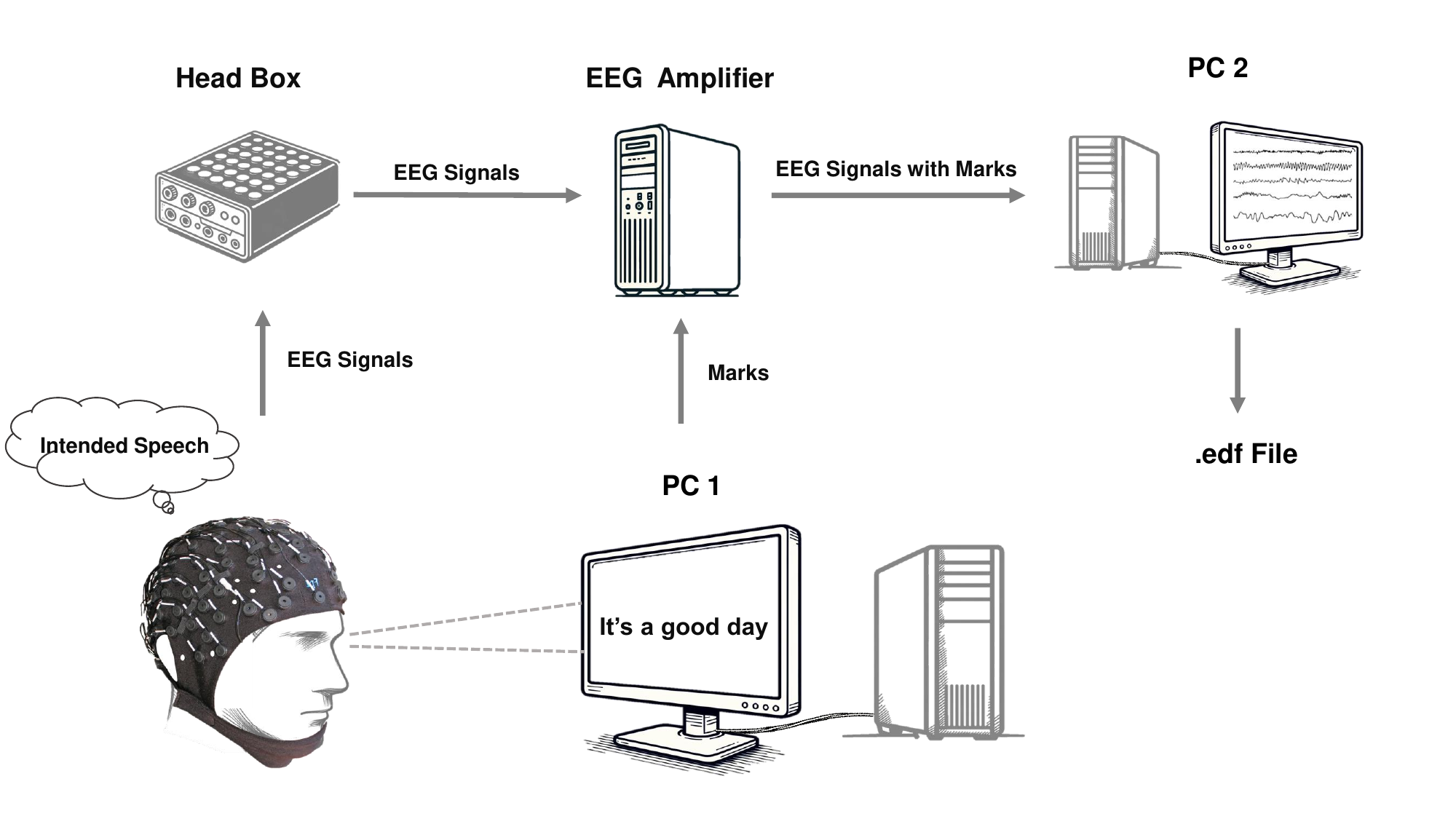}
  \caption {Main experimental setup.}
  \label{sp1}
\end{figure*}

To familiarize the participant with the experimental procedure and the room environment, a detailed explanation of all experimental steps was provided during the placement of the EEG cap and external electrodes. This setup process took approximately 30 minutes. Figure \ref{sp1} shows the main experiment setup. During the experiment, stimuli were presented to the subjects via a screen connected to PC1. The EEG cap recorded the signals, which were filtered and amplified by the headbox. These signals were further amplified by an external amplifier and tagged with markers generated by PC1 to enhance signal integrity. The marker signals enabled accurate segmentation of the EEG data. The processed EEG signals were then transmitted to the experimenter’s computer (PC2) for storage, with the experimental operator monitoring the session through PC2. To minimize interference with the subject, only the PC1 screen was placed inside the data collection room, while all other equipment was located in the control room.

\newpage

\section{EEG Data Preprocessing Pipeline}
\label{EEG Data Preprocessing Pipeline}

\begin{figure*}[t]
  \includegraphics[width=\linewidth, page=2]{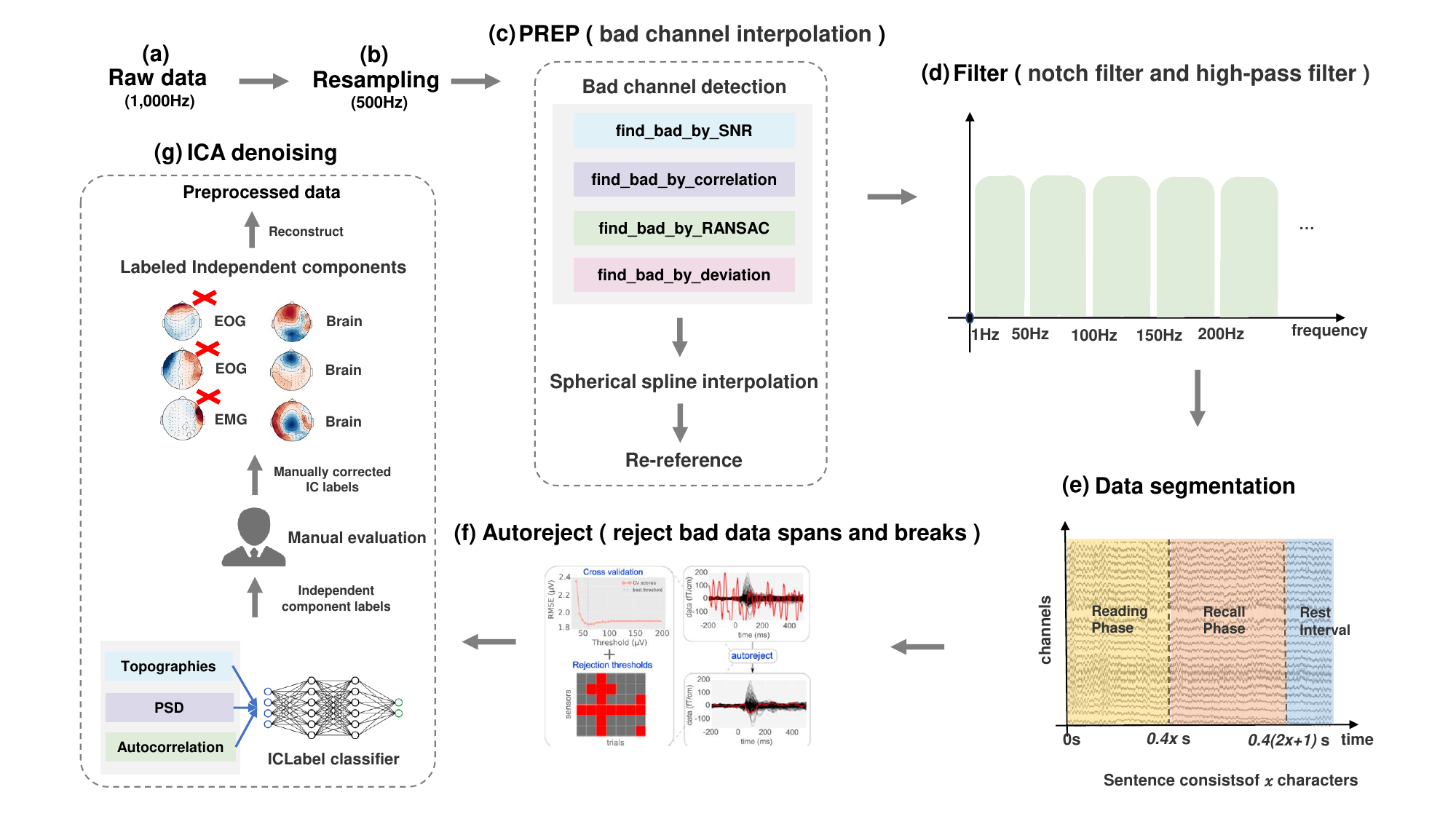}
  \caption {EEG preprocessing pipeline (part (f) cites previous work\cite{jas2017autoreject}). (a) Raw data were recorded at a sampling rate of 1,000 Hz, stored in the original .edf files. (b) The raw data were then downsampled to 500 Hz. (c) The PREP algorithm was applied for bad channel detection and re-referencing. (d) Power line noise was removed, and high-pass filtering was performed. (e) The data were segmented according to the experimental paradigm. (f) Bad data spans and breaks were rejected using the Autoreject algorithm. (g) ICA algorithm was used to remove noise such as EOG and EMG.}
  \label{sp2}
\end{figure*}

Raw EEG data were initially recorded at a sampling rate of 1,000 Hz and stored in the original .edf format. Subsequently, the data were downsampled to 500 Hz to facilitate processing. The PREP algorithm was then applied to detect and remove bad channels and to perform re-referencing. Following this, power line noise was eliminated, and a high-pass filter was applied to further enhance signal quality. The data were then segmented according to the experimental paradigm. Bad data spans and artifacts were identified and rejected using the Autoreject algorithm, as previously described by Jas et al. (2017). Finally, independent component analysis (ICA) was employed to remove physiological noise sources such as electrooculogram (EOG) and electromyogram (EMG) artifacts, resulting in a clean dataset ready for further analysis. The complete process is shown in the Figure \ref{sp2}.

\end{document}